\documentclass[10pt,twocolumn,letterpaper]{article}

\usepackage{iccv}
\usepackage{times}
\usepackage{epsfig}
\usepackage{graphicx}
\usepackage{amsmath}
\usepackage{amssymb}
\usepackage{mathtools}

\usepackage{microtype,graphicx,subfigure,booktabs,bbm,enumitem}
\usepackage{multibib}
\usepackage{multirow}
\usepackage{algorithm}
\usepackage{algorithmic}
\usepackage{pifont}
\newcommand{\cmark}{\ding{51}}%
\newcommand{\xmark}{\ding{55}}%
\newcommand{\stdv}[1]{\scriptsize$\pm$#1}

\usepackage[pagebackref=true,breaklinks=true,colorlinks,bookmarks=false]{hyperref}

\iccvfinalcopy 

\def\iccvPaperID{9812} 

\ificcvfinal\fi
\newcites{SM}{References}
\begin{document}

\title{Co$^2$L: Contrastive Continual Learning}

\author{Hyuntak Cha \qquad Jaeho Lee \qquad Jinwoo Shin \\
Korea Advanced Institute of Science and Technology (KAIST)\\
Daejeon, South Korea\\
{\tt\small \{hyuntak.cha, jaeho-lee, jinwoos\}@kaist.ac.kr}

}

\maketitle
\ificcvfinal\thispagestyle{empty}\fi

\begin{abstract}
Recent breakthroughs in self-supervised learning show that such algorithms learn visual representations that can be transferred better to unseen tasks than joint-training methods relying on task-specific supervision.
In this paper, we found that the similar holds in the \textit{continual learning} context: contrastively learned representations are more robust against the catastrophic forgetting than jointly trained representations. 
Based on this novel observation, we propose a rehearsal-based continual learning algorithm that focuses on continually learning and maintaining transferable representations. 
More specifically, the proposed scheme (1) \textit{learns} representations using the contrastive learning objective, and (2) \textit{preserves} learned representations using a self-supervised distillation step.
We conduct extensive experimental validations under popular benchmark image classification datasets, where our method sets the new state-of-the-art performance.

\end{abstract}

\vspace{-2em}
\section{Introduction}
\begin{figure*}
\vspace{-1em}
\begin{center}
\centerline{\includegraphics[width=0.9\textwidth]{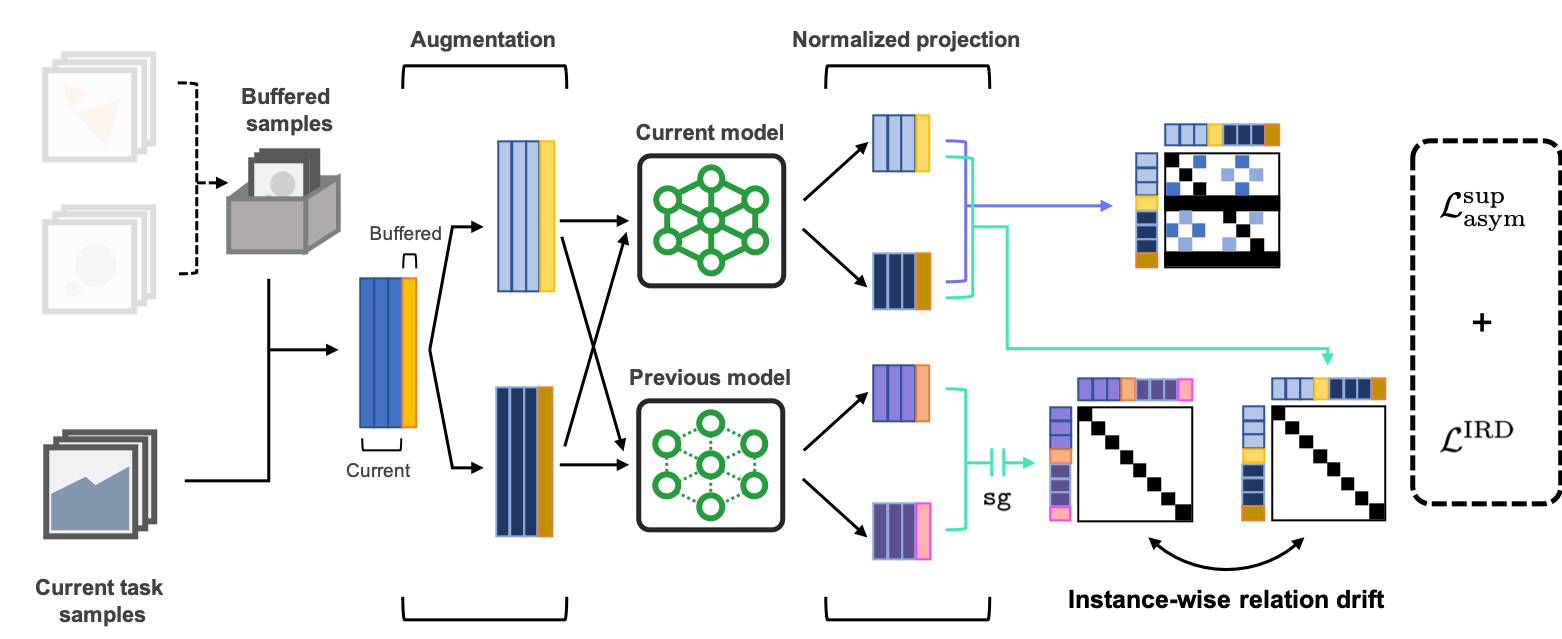}}
\caption{An overview of the Co$^2$L framework. Mini-batch samples from the current task and the memory buffer are augmented and passed through current and past (stored at the end of the previous task) representations. Co$^2$L minimizes the weighted sum of two losses: (1) Asymmetric SupCon loss contrasts anchor samples from the current task against the samples from other classes (Section~\ref{subsection:asymsup}). (2) IRD loss measures the drift of the instance-wise similarities given by the current model from the one given by the previous model (Section~\ref{subsection:IRD}).}
\label{fig:mainfig}
\end{center}
\vspace{-2.5em}
\end{figure*}

Modern deep learning algorithms show impressive performances on the task at hand, but it is well known that they often struggle to retain their knowledge on previously learned tasks after being trained on a new one \cite{mccloskey89}. To mitigate such ``catastrophic forgetting,'' prior works in the continual learning literature focus on \textit{preserving} the previously learned knowledge using various types of information about the past task. Replay-based approaches store a small portion of past samples and rehearse the samples along with present task samples \cite{robins95,lopezpaz17,riemer19,buzzega20}. Regularization-based approaches force the current model to be sufficiently close to the past model---which may be informative about the past task---in the parameter/functional space distance \cite{kirkpatrick17,chaudhry18,serra18}. Expansion-based approaches allocate a unit (\eg, network node, sub-network) for each task and keep the unit untouched during the training for other tasks \cite{rusu16,mallya18}.


In this paper, instead of asking how to isolate previous knowledge from new knowledge, we draw attention to the following fundamental question:
\vspace{0.1in}
\begin{quotation}\em
What type of knowledge is likely to be useful for future tasks (and thus not get forgotten), and how can we learn and preserve such knowledge?
\end{quotation}
\vspace{-0.1in}
To demonstrate its significance, consider the simple scenario that the task at hand is to classify the given image as an apple or a banana. An easy way to solve this problem is to extract and use the color feature of the image; red means apple, and yellow means banana. The color, however, will no longer be useful if our future task is to classify another set of images as apples or strawberries; color may not be used anymore and eventually get forgotten. On the other hand, if the model had learned more complicated features, \eg, shape/polish/texture, the features may be re-used for future tasks and remain unforgotten. This line of thoughts suggests that forgetting does not only come from the limited access to the past experience, but also from the innately restricted access to future events; to suffer less from forgetting, learning more \textit{transferable representations} in the first hand may be as important as carefully preserving the knowledge gained in the past.


To learn more transferable representations for continual learning, we draw inspirations from a recent advance in self-supervised learning, in particular, \textit{contrastive learning} \cite{hadsell06,chen20}. Contrastive methods learn representations using the inductive bias that the prediction should be invariant to certain input transformations instead of relying on task-specific supervisions. 
Despite their simplicity, such methods are known to be surprisingly effective;
for ImageNet classification \cite{ILSVRC15}, contrastively trained representations closely achieve the fully-supervised performance even without labels \cite{chen20} and outperform joint-trained counterparts in the supervised case \cite{khosla20}.
More importantly, while the methods are originally proposed for better in-domain\footnote{The term `in-domain' is used here for the setup where data distributions for representation learning and linear classifier training are the same.} 
performance, recent works also show that such methods provide significant performance gains on unseen domains \cite{chen20,he20}.
Under a continual scenario, we make a similar observation: 
\textit{contrastively learned representations suffer less from forgetting than the jointly trained ones} (see Section~\ref{subsection:observation} for details).

Unfortunately, applying this idea to continual settings is not straightforward due to at least two reasons: First, having access to informative negative samples is known to be crucial for the success of contrastive learning \cite{robinson21}, while the instantaneous demographics of negatives samples are severely restricted under standard continual setups; in class-incremental learning, for instance, it is common to assume that the learner can access samples from only a small number of classes at each time step. 
Second, the question of how to preserve the learned \textit{representations} not as a part of a jointly trained representation-classifier pair has not been fully answered. Indeed, recent works on representation learning for continual setups aim to learn representations accelerating future learning under a similar decoupled learning setup but lack an explicit design to preserve representations. 


\noindent\textbf{Contribution.} To address these challenges, we propose a new rehearsal-based continual learning algorithm, coined \textbf{Co$^2$L} (\textbf{Co}ntrastive \textbf{Co}ntinual \textbf{L}earning).
Unlike previous continual (representation) learning methods, we aim to \textit{learn} and \textit{preserve} representations continually in a decoupled representation-classifier scheme. The overview of Co$^2$L is illustrated in Figure~\ref{fig:mainfig}.

Our contribution under this setup is twofold:
\begin{enumerate}[leftmargin=*,topsep=0pt,itemsep=-0.5ex,partopsep=1ex,parsep=1ex]
\item \textit{Contrastive learning:} 
We design an asymmetric version of supervised contrastive loss for learning representations under continual learning setup (Section~\ref{subsection:asymsup}) and empirically show its benefits on improving the representation quality. 
\item \textit{Preserving representations:} 
We propose a novel preservation mechanism for contrastively learned representations, which works by self-distillation of instance-wise relations (Section~\ref{subsection:IRD}); to the best of our knowledge, this is a first method explicitly designed to preserve representations without a jointly trained classifier. 
\end{enumerate}
We validate Co$^2$L under various experimental scenarios encompassing task-incremental learning, domain-incremental learning, and class-incremental learning. Co$^2$L consistently outperforms all baselines on various datasets, scenarios, and memory setups.
With careful ablation studies, we also show that both components we propose (asymmetric supervised contrastive loss, instance-wise relation distillation) are essential for performance.
In the ablation of distillation, we empirically show that distillation preserves learned representations and efficiently uses buffered samples, which might be the main source of consistent gains over all comparisons: distillation provides 22.40\% and 10.59\% relative improvements with/without buffered samples respectively on the Seq-CIFAR-10 dataset. 
In the ablation of asymmetric supervised contrastive loss, we quantitatively verify that the asymmetric version consistently provides performance gains over the original one on all setups, \eg, 8.15\% relative improvement on the Seq-CIFAR-10 with buffer size 500. We also provide qualitative implication on this performance gain by visualizing learned representations, which shows our asymmetric version prevents severe drifts of learned features.

\vspace{-0.3em}
\section{Related Work}
\label{section:related-work}
\noindent\textbf{Rehearsal-based continual learning.}
Continual learning methods have been developed in three major streams: using a fixed-sized buffer to replay past samples (rehearsal-based approach), regulating model parameter changes through learning (regularization-based approach), or dynamically expanding model architecture on demand (expansion-based approach). Among them, the rehearsal-based approach has shown great performance in continual learning settings, albeit with its simplicity. The idea of Experience Replay (ER \cite{riemer19}) is simply managing a fixed-sized buffer to retain a small number of samples and replaying those samples to prevent forgetting past knowledge. Following this simple setup, several methods have been proposed by expanding this framework in two aspects: which samples should be stored and how to utilize stored samples. 
Those works mainly focus on either regulating model updates not to contradict the learning objectives on past samples \cite{lopezpaz17, buzzega20} or selecting the most representative/forgetting-prone samples to prevent changes in past predictions \cite{aljundi19, chaudhry20, Rebuffi17}. In a purely decoupled representation learning setup, however, there are few studies on those two aspects since learning objectives of representation learning may not be directly aligned to the task-specific objectives in typical joint training schemes. In this work, we focus on utilizing buffered samples to learn representations continually on a decoupled representation-classifier learning scheme.

\noindent\textbf{Representation learning in continual learning.}
Only a few recent studies on continual learning focus on representations models learned in two aspects: how to maintain learned representations \cite{Rebuffi17} and how to learn representations accelerating future learning \cite{javed19, gupta20}. iCaRL \cite{Rebuffi17} prevents representations from being forgotten by leveraging distillation.
Recent approaches \cite{javed19, gupta20} leverage meta-learning \cite{finn17} separate concerns of representation learning and classifier training and directly optimizes objectives that minimize forgetting by learning representations that accelerate future learning on meta-learning frameworks. 
In this work, we focus on constructing representations that suffer less from forgetting and also focus on preserving learned representations in continual learning context not as a part of joint training. 

\noindent\textbf{Contrastive representation learning.}
Recent progress in contrastive representation learning shows superior downstream task performance, even competitive to supervised training. Noise-contrastive estimation \cite{gutmann10} is the seminal work that estimates the latent distribution by contrasting with artificial noises. Info-NCE \cite{oord18} tries to learn representations from visual inputs by leveraging an auto-regressive model to predict the future in an unsupervised manner. Recent advances in this area stem from the use of multiple views as positive samples \cite{tian19}.
These core concepts have been followed by studies \cite{chen20, he20, grill2020bootstrap, chen2020exploring} that have resolved practical limitations that have previously made learning difficult such as negative sample pairs, large batch size, and momentum encoders. Meanwhile, it has been shown that supervised learning can also enjoy the benefits of contrastive representation learning by simply using labels to extend the definition of positive samples \cite{khosla20}. 
In this work, we mainly leverage contrastive representation learning schemes on the continual learning setup based on our novel observation (Section~\ref{subsection:observation}).

\vspace{0.2em}
\noindent\textbf{Knowledge distillation}. 
In continual learning, knowledge distillation is widely used to mitigate forgetting by distilling past signatures to the current models \cite{Li16, Rebuffi17}. However, it has not been studied to design/utilize knowledge distillation for decoupled representation-classifier training in the continual learning setup. In this work, we develop novel self-distillation loss for contrastive continual learning, which is inspired by the recently proposed distillation loss \cite{fang20} for contrastive learning framework.


\vspace{-0.3em}
\section{Problem Setup and Preliminaries}
\label{section:background}
In this section, we formalize the considered continual learning setup and briefly describe a recently proposed supervised contrastive learning scheme \cite{khosla20} that will be used as the main framework for designing Co$^2$L (Section~\ref{section:method}).
\subsection{Problem Setup: Continual Learning}
\label{subsection:problem-setup}

We consider three popular scenarios of continual learning as categorized by \cite{vandeven2019three}: task-incremental learning (Task-IL), domain-incremental learning (Domain-IL), and class-incremental learning (Class-IL).

Formally, the learner is trained on a sequence of tasks indexed by $t \in \{1,2,\ldots,T\}$. For each task, we suppose that there is a task-specific class set $C_t$. For Task-IL and Class-IL, $\{C_t\}_{t=1}^T$ are assumed to be disjoint, \ie,
\begin{align}
    t \ne t' \Rightarrow C_t \cap C_{t'} = \emptyset, \qquad \text{(Task/Class-IL).}
\end{align}
For Domain-IL, $C_t$ remains the same throughout the tasks:
\begin{align}
    C_1 = C_2 = \cdots = C_T, \qquad \text{(Domain-IL)}.
\end{align}
During each task, $n_t$ copies of training input-label pairs are independently drawn from some task-specific distribution, \ie, $\{(\mathbf{x}_i,y_i)\}_{i=1}^{n_t} \sim D_t$.
Here, $\mathbf{x}$ denotes the input image, and $y_i \in C_t$ denotes the class label belonging to the task-specific class set. For Task-IL, the learned models are assumed to have access to the task label $t$ during the test phase; the goal is to find a predictor $\varphi_{\theta}(\mathbf{x},t)$ parameterized by $\theta$ such that
\begin{align}
    \mathcal{L}(\theta) := \sum_{t=1}^T \mathbb{E}_{D_t}[\ell(y,\varphi_\theta(x,t))], \qquad \text{(Task-IL)}
\end{align}
is minimized for some loss function $\ell(\cdot,\cdot)$. For Domain-IL and Class-IL, the model cannot access the task label during the test phase; the goal is to find a predictor $\varphi_{\theta}(\mathbf{x})$ minimizing
\begin{align}
    \mathcal{L}(\theta) = \sum_{t=1}^T \mathbb{E}_{D_t}[\ell(y,\varphi_\theta(x))], \quad \text{(Domain/Class-IL).}
\end{align}

\begin{figure*}[t]
\vspace{-1em}
\centering
\subfigure[Asymmetric SupCon Loss]{
\centering
\includegraphics[height=1.8in]{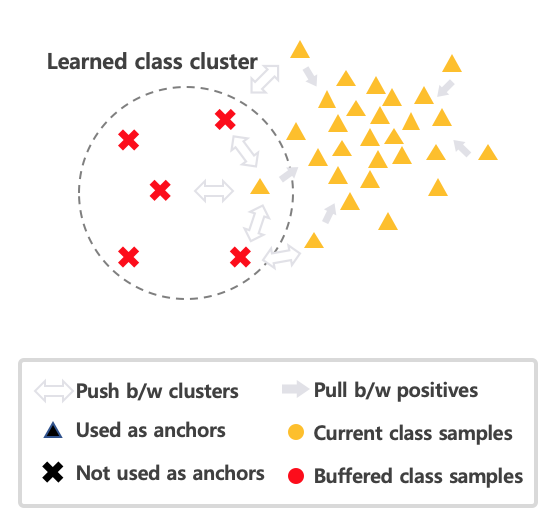}
\label{subfig:asymsup}
}
\subfigure[Instance-wise Relation Distillation Loss]{
\centering
\includegraphics[height=1.8in]{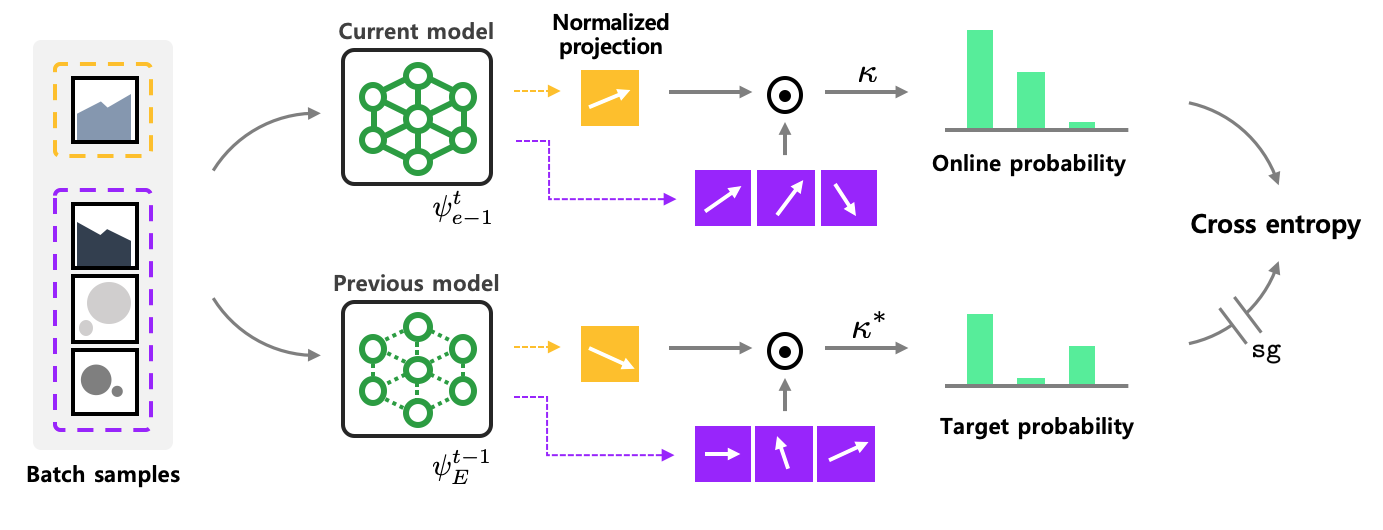}
\label{subfig:ird}
}
\caption{
Illustration of Asymmetric Supervised Constrastive Loss and Instance-wise Relation Distillation (IRD). (a) Given augmented mini-batch samples, asymmetric SupCon considers samples from the same class of the current task as positives. In other words, the pulling effects between anchors only exist between current task samples. (b) Given augmented mini-batch samples, the instance-wise relation is defined on the normalized projected feature vectors. The relation vectors, \ie, dot products ($\odot$) of feature vectors, are computed from the learnable ($\psi^t_{e-1}$) and reference model ($\psi^{t-1}_{E}$), respectively. For $E$ epoch training, such temperature scaled relation is distilled from the reference model to the learnable model. Note that the reference model is snapped at the end of $(t-1)$\textit{-th} task training, and we only update the learnable model's weights using stop-gradient (denoted by $\mathtt{sg}$).}
\label{fig:submethods}
\vspace{-0.5em}
\end{figure*}
\subsection{Preliminaries: Contrastive Learning}
We now describe the SupCon (Supervised Contrastive learning) algorithm, proposed by \cite{khosla20}. Suppose that the classification model can be decomposed into two components 
\begin{align}
    \varphi_\theta = \mathbf{w} \circ f_{\vartheta}
\end{align}
with parameter pairs $\theta = (\vartheta,\mathbf{w})$, where $\mathbf{w}(\cdot)$ is the linear classifier and $f_{\vartheta}(\cdot)$ is the representation. Without training $\mathbf{w}$, SupCon directly trains $f_{\vartheta}$ as follows: Given a batch of $N$ training samples $\{(\mathbf{x}_i,y_i)\}_{i=1}^N$, SupCon first generates an augmented batch $\{(\tilde{\mathbf{x}}_i,\tilde{y}_i)\}_{i=1}^{2N}$ by making a two randomly augmented versions of $\mathbf{x}_k$ as $\tilde{\mathbf{x}}_{2k}, \tilde{\mathbf{x}}_{2k-1}$, with $\tilde{y}_{2k} = \tilde{y}_{2k-1} = y_k$. The samples in the augmented batch are mapped to a unit $d$-dimensional Euclidean sphere as
\begin{align}
    \mathbf{z}_i = (g \circ f)_{\psi}(\tilde{\mathbf{x}}_i),
\end{align}
where $g = g_{\phi}$ denotes the projection map parametrized by $\phi$, and $\psi$ denotes the concatenation of $\vartheta$ and $\phi$. Now, the feature map $(g \circ f)_{\psi}$ is trained to minimize the supervised contrastive loss
\begin{align}
    \mathcal{L}^{\text{sup}} = \sum_{i = 1}^{2N} \frac{-1}{|\mathfrak{p}_i|}\sum_{j \in \mathfrak{p}_i} \log \left(\frac{\exp(\mathbf{z}_i \cdot \mathbf{z}_j / \tau)}{\sum_{k \ne i}\exp(\mathbf{z}_i \cdot \mathbf{z}_k / \tau)}\right), \label{eqn:supcon}
\end{align}
where $\tau > 0$ is some temperature hyperparameter and $\mathfrak{p}_i$ is the index set of positive samples with respect to the anchor $\tilde{\mathbf{x}}_i$, defined as
\begin{align}
    \mathfrak{p}_i = \big\{j \in \{1,\ldots,2N\} ~\big|~\:j \ne i,\: y_j = y_i\big\}.
\end{align}
In other words, the sample in $\mathfrak{p}_i$ is either the other augmentation of the unaugmented version of $\tilde{\mathbf{x}}_i$, or one of the other augmented samples having the same label.

\vspace{-0.3em}
\section{Co$^2$L: Contrastive Continual Learning}

\label{section:method}
Here, we propose a rehearsal-based contrastive continual learning scheme, coined \textbf{Co$^2$L} (\textbf{Co}ntrastive \textbf{Co}ntinual \textbf{L}earning). At a high level, Co$^2$L (1) \textit{learns} the representations with an asymmetric form of supervised contrastive loss (Section~\ref{subsection:asymsup}) and (2) \textit{preserves} learned representations using self-supervised distillation (Section~\ref{subsection:IRD}) in a decoupled representation-classifier training scheme. This is done by a mini-batch gradient descent based on the compound loss 
\begin{align}
\label{eqn:total-loss}
    &\mathcal{L} = \underbrace{\mathcal{L}^\text{sup}_\text{asym}}_\text{(1) learning} + 
    \underbrace{\vphantom{\mathcal{L}^\text{sup}_\text{asym}}\lambda\cdot \mathcal{L}^{\text{IRD}}}_{\text{(2) preserving}}.
\end{align}
Here, each batch is composed of two independently augmented views of $N$ samples (thus $2N$ in total), where each sample is drawn from the union of current task samples and buffered samples.

\subsection{Representation Learning with Asymmetric Supervised Contrastive Loss}
\label{subsection:asymsup}

For learning representation continually, we use an asymmetrically modified version of the SupCon objective $\mathcal{L}^{\text{sup}}$. In the modified version, we only use current task samples as anchors; past task samples from the memory buffer will only be used as negative samples (see Figure~\ref{subfig:asymsup}). Formally, if we let $S \subset \{1,\ldots,2N\}$ be the set of indices of current task samples in the batch, the modified supervised contrastive loss is defined as
\begin{align}
    \label{eqn:supcon_asym}
    \mathcal{L}^{\text {sup }}_{\text{asym}} &=\sum_{i \in S}  \frac{-1}{|\mathfrak{p}_i|} \sum_{p \in \mathfrak{p}_i} \log \frac{\exp \left(\mathbf{z}_{i} \cdot \mathbf{z}_{p} / \tau\right)}{\sum_{k \ne i} \exp \left(\mathbf{z}_{i} \cdot \mathbf{z}_{k} / \tau\right)}.
\end{align}
The motivation behind this asymmetric design is to prevent a model from overfitting to a small number of past task samples. It turns out that such a design indeed helps to boost the performance. In Section~\ref{subsection:ablation}, we empirically observe that the asymmetric version $\mathcal{L}^{\text{sup }}_{\text{asym}}$ outperforms the original $\mathcal{L}^{\text{sup}}$ and generates better-spread features of buffered samples.

\subsection{Instance-wise Relation Distillation (IRD) for Contrastive Continual Learning}
\label{subsection:IRD}

While using the contrastive learning objective (eq.~\ref{eqn:supcon_asym}) readily provides a more transferable representation, one may still benefit from having an explicit mechanism to preserve the learned knowledge. Taking the inspiration from \cite{fang20}, we propose an instance-wise relation distillation (IRD); IRD regulates the changes in feature relation between batch samples via self-distillation  (see Figure~\ref{subfig:ird}).
Formally, we define the IRD loss $\mathcal{L}^{\text{IRD}}$ as follows: For each sample $\tilde{\mathbf{x}}_i$ in a batch $\mathcal{B}$, we define the \textit{instance-wise similarity vector}
\begin{align}
\label{eqn:simvec}
    \mathbf{p}\left(\tilde{\mathbf{x}}_{i} ; \psi, \kappa\right)=\left[p_{i, 1},  \ldots, p_{i, i-1}, p_{i, i+1},\ldots,p_{i,2N}\right],
\end{align}
where $p_{i,j}$ denotes the normalized instance-wise similarity
\begin{align}
p_{i,j}&=\frac{\exp \left(\mathbf{z}_{i} \cdot \mathbf{z}_{j} / \kappa\right)}{\sum_{k \neq i}^{2N} \exp \left(\mathbf{z}_{i} \cdot \mathbf{z}_{k} / \kappa\right)}
\end{align}
given the representation parameterized by $\psi$ and the temperature hyperparameter $\kappa$. In other words, the instance-wise similarity vector $\mathbf{p}(\cdot)$ is the normalized similarity of a sample to other samples in the batch.

Roughly, the IRD loss quantifies the discrepancy between the instance-wise similarities of the current representation and the past representation; the past representation is a snapshot of the model at the end of the previous task. Denoting the parameters of the past/current model as $\psi^{\text{past}}$ and $\psi$, the IRD loss is defined as
\begin{align}
\label{eqn:IRD}
    \mathcal{L}^{\text {IRD }}  &=\sum_{i=1}^{2N}-\mathbf{p}\left(\tilde{\mathbf{x}}_{i} ; \psi^{\text{past}}, \kappa^{\ast}\right) \cdot \log \mathbf{p}\left(\tilde{\mathbf{x}}_{i} ; \psi, \kappa\right),
\end{align}
where the logarithms and multiplications on the vectors denote the entrywise logarithms and multiplications. We note that we are using different temperature hyperparameters for the past and current similarity vectors; on the other hand, both $\kappa,\kappa^*$ will remain fixed throughout the tasks.

By using fixed model weights snapped at the end of previous task training as the reference model $\psi^{\text{past}}$, IRD distills learned representations to the current training model $\psi$, thereby leading to preserving learned representations. Since contrastive representation learning stems from deep metric learning, IRD achieves knowledge preservation by regulating overall \textit{structure} changes of learned representations. Note that IRD does not regulate exact changes in feature space and does not define relation from encoder outputs like \cite{fang20}. More detailed comparisons between \cite{fang20} and ours are provided in the supplementary material.

\begin{figure*}
\begin{center}
\centerline{\includegraphics[width=\textwidth]{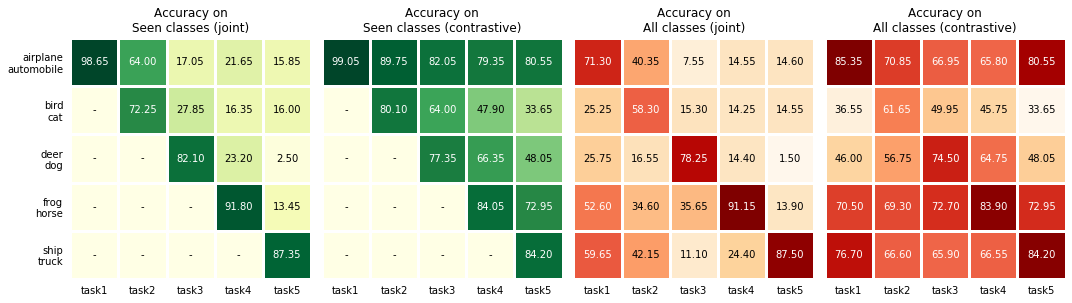}}
\caption{Observation on two learning schemes, representation-classifier joint training and contrastive representation learning on Seq-CIFAR-10 without any design used for the continual learning settings. As new task arrives, each model is trained only with current task samples with model weights without re-initialization. After each task training ends, a new linear classifier is trained on the fixed current representation with samples observed so far (denoted by ``seen objects'') or all samples including ones from future tasks (denoted by ``all objects''). The pair of left figures shows contrastively trained representations suffer less from forgetting than the joint trained ones. The right pair shows contrastively learned representation is much more useful to perform unseen objects classification tasks.}
\label{fig:observation}
\end{center}
\vspace{-2em}
\end{figure*}

\subsection{Algorithm Details}
\label{subsection:Co$^2$L}

\begin{algorithm}[t]
\caption{Co$^2$L: Contrastive Continual Learning}
\label{alg:whole}
\begin{algorithmic}[1]

\STATE \textbf{Input}: 
Encoder parameters $\vartheta$, 
projector parameters $\phi$, 
number of tasks $T$, 
family of augmentations $\mathcal{H}$,
a set of training sets $\{\{(x^t_i, y^t_i)\}\}^T_{t=1}$, 
a set of disjoint class sets $\{\mathcal{C}_t\}^T_{t=1}$, 
learning rate $\eta$, 
number of epochs of $t$-th task $E_t$, 
distillation temperatures $\kappa,\kappa^{\ast}$, 
distillation power $\lambda$.

\STATE Initialize network $(g \circ f)_{\psi} (\cdot)$ where $\psi = (\vartheta,\phi)$.
\FOR{$t = 1, \cdots, T$}

\STATE Construct dataset $\mathcal{D}$ by $\mathcal{D} \gets \{(x^t_i, y^t_i)\} \cup \mathcal{M}$\\


\FOR{$e = 1, \cdots, E_t$}

\STATE Draw a mini-batch $\{(x_i, y_i)\}_{i=1}^{N}$ from $\mathcal{D}$\\

\FORALL{$k \in \{1, \cdots ,N\}$}
\STATE Draw two augmentations $h\sim\mathcal{H}, h'\sim\mathcal{H}$ \\
\STATE Initialize 
anchor indices sets $S \gets \emptyset, I \gets \emptyset$
\STATE $\tilde{x}_{2k-1} = h(x_k)$\\
\STATE $\tilde{x}_{2k} = h'(x_k)$\\
\STATE $I \gets I \cup \{2k-1, 2k\}$
\IF {$y_k \in C_t$}
\STATE $S \gets S \cup \{2k-1, 2k\}$
\ENDIF
\ENDFOR

\STATE Compute $\mathcal{L}$ by $\mathcal{L} \gets \mathcal{L}^\text{sup}_\text{asym}(I, S; \psi^{t}_{e-1})$ (eq.~\ref{eqn:supcon_asym})\\
\IF {$t > 1$}
\STATE Update $\mathcal{L}$ by \\ $\mathcal{L} \gets \mathcal{L} + \lambda \cdot \mathcal{L}^\text{IRD}(\psi^{t-1}_{E_{t-1}}, \psi^{t}_{e-1}, \kappa^{\ast}, \kappa)$ (eq.~\ref{eqn:IRD})
\ENDIF

\STATE Update 
$\psi^t_{e-1}$
by $\psi^t_{e} \gets \psi^t_{e-1} - \eta \nabla_{\psi^t_{e-1}} \mathcal{L}$ \\
\ENDFOR

\STATE Manage buffer $\mathcal{M}$ for the number of each class samples to be same by uniform sampling.
\ENDFOR
\end{algorithmic}
\end{algorithm}
\vspace{-0.5em}
Here, we give a complete picture of the overall training procedure and give additional details. The full algorithm is provided in Algorithm \ref{alg:whole}. 

\vspace{0.2em}
\noindent\textbf{Data preparation.} 
As the initial or new task arrives, the dataset is built as a union of current task samples and buffered samples, without any oversampling \cite{Chawla02, han05}. The mini-batch is drawn from this dataset, where each sample is independently drawn with equal probability. To enjoy the benefits of contrastive representation learning, each sample is augmented into two views following \cite{Chen19}. The detailed augmentation scheme for contrastive learning is provided in the supplementary material.

\vspace{0.2em}
\noindent\textbf{Learning new representation.} The augmented samples are forwarded to the encoder $f_{\vartheta}$ and projection map $g_{\phi}$ sequentially. The projection map outputs are used to compute asymmetric supervised contrastive loss (eq.~\ref{eqn:supcon_asym}).

\vspace{0.2em}
\noindent\textbf{Preserving learned representation.}
When a new task arrives (\ie, $t > 1$), we compute instance-wise relation drifts between reference model and the training model with IRD loss (eq.~\ref{eqn:IRD}). To this end, we settle the reference model as the trained model at the end of the training of $(t-1)$-th task. Note that while optimizing total loss (eq.~\ref{eqn:total-loss}), the reference model is not updated.

\vspace{0.2em}
\noindent\textbf{Buffer management.}
At the end of training each task, a small portion of training samples is pushed into a replay buffer. Due to its buffer size constraint, a small subset of samples from each class is pulled out of the replay buffer at the same ratio. The sample to be pushed or pulled is uniformly randomly selected for all procedures.

\vspace{-0.3em}
\section{Experiment}

\label{section:experiment}
\subsection{Experimental Setup}
\label{subsection:exp_setup}
\noindent\textbf{Learning scenarios and datasets.}
Following \cite{vandeven2019three}, we conduct continual learning experiments on Task Incremental Learning (Task-IL), Class Incremental Learning (Class-IL) and Domain Incremental Learning (Domain-IL) scenarios. 
We conduct experiments on Seq-CIFAR-10 and Seq-Tiny-ImageNet for Task-IL and Class-IL scenarios and R-MNIST for Domain-IL scenario. 
\textbf{Seq-CIFAR-10} is the set of splits (tasks) of the CIFAR-10 \cite{krizhevsky2009learning} dataset. We split the CIFAR-10 dataset into five separate sample sets, and each sample set consists of two classes. Similarly, \textbf{Seq-Tiny-ImageNet} is built from Tiny-ImageNet \cite{Le2015TinyIV} by splitting 200 class samples into 10 disjoint sets of samples, each consisting of 20 classes.
Seq-CIFAR-10 and Seq-Tiny-ImageNet split are given in the same order across different runs, as in \cite{buzzega20}.
We conduct experiments on \textbf{R-MNIST} \cite{lopezpaz17} for Domain-IL experiments. 
For Domain-IL scenario, R-MNIST is constructed by rotating the original MNIST \cite{lecun98} images by a random degree in the range of [0, $\pi$). R-MNIST consists of 20 tasks, corresponding to 20 uniformly randomly chosen degrees.
We note that we treat samples from different domains with the same digit class as different classes while applying asymmetric supervised contrastive loss. 


\vspace{0.2em}
\noindent\textbf{Training.}
We compare our contrastive continual learning algorithm with rehearsal-based continual learning baselines: ER \cite{riemer19}, iCaRL \cite{Rebuffi17}, GEM \cite{lopezpaz17}, A-GEM \cite{chaudhry19}, FDR \cite{benjamin19}, GSS \cite{aljundi19}, HAL \cite{chaudhry20}, DER \cite{buzzega20}, and DER++ \cite{buzzega20}. We train ResNet-18 \cite{He2015} on Seq-CIFAR-10 and Tiny-ImageNet, and a simple network with convolution layers on R-MNIST. 
For all baselines, we report performance given in \cite{buzzega20} of buffer size 200 and 500 except for R-MNIST since we choose a different architecture. 
More training details are provided in the supplementary material.


\vspace{0.2em}
\noindent\textbf{Evaluation protocol for Co$^2$L.}
As Co$^2$L is a representation learning scheme and not a joint representation-classifier training, we need to train a classifier additionally. For a fair comparison, we train a classifier using only the last task samples and buffered samples on top of the frozen representations learned by Co$^2$L. To avoid the class-imbalance problems, we train a linear classifier with a class balanced sampling strategy, where first a class is selected uniformly from the set of classes, and then an instance from that class is subsequently uniformly sampled. 
We train a linear classifier for 100 epochs for all experiments, and we report classification test accuracy on this classifier.

\begin{table*}[ht]
\setlength{\tabcolsep}{0.2cm}
\centering
\begin{center}
\begin{small}
\resizebox{0.70\textwidth}{!}{
\begin{tabular}{clccccccccccc}
\toprule
\multirow{2}{*}{ Buffer } & Dataset & & \multicolumn{2}{c}{ Seq-CIFAR-10 } & & \multicolumn{2}{c}{ Seq-Tiny-ImageNet } & & R-MNIST & \\
\cmidrule{4-5} \cmidrule{7-8} \cmidrule{10-10}
& Scenario & & Class-IL & Task-IL  & & Class-IL & Task-IL & & Domain-IL & \\
\midrule
\multirow{10}{*}{ 200 } 
& ER \cite{riemer19}
& & 44.79\stdv{1.86} & 91.19\stdv{0.94} 
& & 8.49\stdv{0.16} & 38.17\stdv{2.00} 
& & 93.53\stdv{1.15} &  \\
& GEM \cite{lopezpaz17}
& & 25.54\stdv{0.76} & 90.44\stdv{0.94} 
& & -  & - 
& & 89.86\stdv{1.23} & \\
& A-GEM \cite{chaudhry19}
& & 20.04\stdv{0.34} & 83.88\stdv{1.49} 
& & 8.07\stdv{0.08} & 22.77\stdv{0.03} 
& & 89.03\stdv{2.76} & \\
& iCaRL \cite{Rebuffi17}
& & 49.02\stdv{3.20} & 88.99\stdv{2.13} 
& & 7.53\stdv{0.79} & 28.19\stdv{1.47} 
& &  - & \\
& FDR \cite{benjamin19}
& & 30.91\stdv{2.74} & 91.01\stdv{0.68} 
& & 8.70\stdv{0.19} & 40.36\stdv{0.68} 
& & 93.71\stdv{1.51} & \\
& GSS \cite{aljundi19}
& & 39.07\stdv{5.59} & 88.80\stdv{2.89} 
& & - & - 
& & 87.10\stdv{7.23} &\\
& HAL \cite{chaudhry20}
& & 32.36\stdv{2.70} & 82.51\stdv{3.20} 
& & - & - 
& & 89.40\stdv{2.50} & \\
& DER \cite{buzzega20}
& & 61.93\stdv{1.79} & 91.40\stdv{0.92} 
& & 11.87\stdv{0.78} & 40.22\stdv{0.67} 
& & 96.43\stdv{0.59} & \\
& DER++ \cite{buzzega20}
& & 64.88\stdv{1.17} & 91.92\stdv{0.60} 
& & 10.96\stdv{1.17} & 40.87\stdv{1.16} 
& & 95.98\stdv{1.06} & \\
& \textbf{Co$^2$L (ours)} 
& & \textbf{65.57\stdv{1.37}}& \textbf{93.43\stdv{0.78}} 
& & \textbf{13.88\stdv{0.40}} & \textbf{42.37\stdv{0.74}} 
& & \textbf{97.90\stdv{1.92}}& \\
\midrule
\multirow{10}{*}{ 500 } 
& ER \cite{riemer19}
& & 57.74\stdv{0.27} & 93.61\stdv{0.27} 
& & 9.99\stdv{0.29} & 48.64\stdv{0.46} 
& & 94.89\stdv{0.95} &  \\
& GEM \cite{lopezpaz17}
& & 26.20\stdv{1.26} & 92.16\stdv{0.64} 
& & - & -
& & 92.55\stdv{0.85} & \\
& A-GEM \cite{chaudhry19}
& & 22.67\stdv{0.57} & 89.48\stdv{1.45} 
& & 8.06\stdv{0.04} & 25.33\stdv{0.49} 
& & 89.04\stdv{7.01} & \\
& iCaRL \cite{Rebuffi17}
& & 47.55\stdv{3.95} & 88.22\stdv{2.62} 
& & 9.38\stdv{1.53} & 31.55\stdv{3.27} 
& &  - & \\
& FDR \cite{benjamin19}
& & 28.71\stdv{3.23} & 93.29\stdv{0.59} 
& & 10.54\stdv{0.21} & 49.88\stdv{0.71} 
& & 95.48\stdv{0.68} & \\
& GSS \cite{aljundi19}
& & 49.73\stdv{4.78} & 91.02\stdv{1.57} 
& & - & - 
& & 89.38\stdv{3.12} &\\
& HAL \cite{chaudhry20}
& & 41.79\stdv{4.46} & 84.54\stdv{2.36} 
& & - & - 
& & 92.35\stdv0.81 & \\
& DER \cite{buzzega20}
& & 70.51\stdv{1.67} & 93.40\stdv{0.39} 
& & 17.75\stdv{1.14} & 51.78\stdv{0.88} 
& & 97.57\stdv{1.47} & \\
& DER++ \cite{buzzega20}
& & 72.70\stdv{1.36} & 93.88\stdv{0.50} 
& & 19.38\stdv{1.41} & 51.91\stdv{0.68} 
& & 97.54\stdv{0.43} & \\
& \textbf{Co$^2$L (ours)} 
& & \textbf{74.26\stdv{0.77}}& \textbf{95.90\stdv{0.26}} 
& & \textbf{20.12\stdv{0.42}} & \textbf{53.04\stdv{0.69}}
& & \textbf{98.65 \stdv{0.31}}& \\
\bottomrule
\end{tabular}
}

\end{small}
\end{center}
\caption{Classification accuracies for Seq-CIFAR-10, Seq-Tiny-ImageNet and R-MNIST on rehearsal-based baselines and our algorithm. We report performance of baslines of Seq-CIFAR-10 and Seq-Tiny-ImageNet from \cite{buzzega20}. ‘-’ indicates experiments unable to run due to compatibility issues (\eg, iCaRL in Domain-IL) or intractable training time (\eg, GEM, HAL or GSS on Tiny ImageNet). All results are averaged over ten independent trials. The best performance marked as bold.
}
\label{main-table}
\vspace{-1em}
\end{table*}

\subsection{Main Results}\label{subsection:observation}


\noindent\textbf{Validation on our key hypothesis.}
Before we provide results of Co$^2$L in comparison with other methods, we first validate our running premise for method design: \textit{Contrastive learning learns more useful representation for the future task than the joint classifier-representation supervised learning.} This premise, however, is not easy to verify under the standard continual learning setup. Indeed, the quality of a representation is typically defined as the joint predictive performance with the best possible (linear) downstream classifier (see, \eg, \cite{arora19}, and references therein), but optimal classifiers are only rarely learned under continual setups.

To circumvent this obstacle, we consider the following synthetic, yet insightful scenario: After training representations under the standard continual setup, we freeze the representations and freshly train the downstream classifier, using training data from \textit{all tasks}. Here, the classifier trained on all observed samples so far will perform learned tasks well unless frozen representations suffer from forgetting.

As shown in the left pair of heatmaps in Figure \ref{fig:observation}, the average test accuracy on the previous tasks is surprisingly higher in \textit{contrastive} than in \textit{joint} (for off-diagonal parts, 21.79\% vs. 66.46\%). In other words, without any specific method to account for continual setup, contrastive method learns representations that suffer less from forgetting than jointly trained ones. 

In the right pair of heatmaps in Figure~\ref{fig:observation}, we report test accuracies of the classifiers that are trained with \textit{all samples}, including the samples from unseen tasks.
Interestingly, we observe that the average task accuracy on the unseen task is also notably higher in contrastively trained representations (rightmost heatmap) than jointly trained ones (second to right); for lower triangle parts, 32.77\% vs. 62.76\%. This implies that contrastive learning methods learn more highly transferable representations to future tasks, which might be the source of its robustness against forgetting.

\vspace{0.2em}
\noindent\textbf{Superiority of Co$^2$L over baselines.}
As shown in Table~\ref{main-table}, our contrastive continual learning algorithm consistently outperforms all baselines in various scenarios, datasets, and memory sizes. 
Such results indicate that our algorithm successfully learns and preserves representations useful for future learning, and thus it significantly mitigates catastrophic forgetting. 
Moreover, such consistent gains over all comparisons show that our scheme is not limited to certain incremental learning scenarios.
In what follows, we provide a more detailed analysis of our algorithm.

\subsection{Ablation Studies}
\label{subsection:ablation}

\noindent\textbf{Effectiveness of IRD.}
To verify the effectiveness of IRD, we perform an ablation experiment with the class-IL setup on the Seq-CIFAR-10 dataset (identical to the setup in Section~\ref{subsection:observation}), with three additional variants of Co$^2$L. \textit{(a) without buffer and IRD:} We optimize using only the SupCon loss (eq.~\ref{eqn:supcon}); the symmetric version is identical to the asymmetric one since we do not use a replay buffer. \textit{(b) with IRD only:} We use both (symmetric) SupCon loss and IRD loss. \textit{(c) with replay buffer only:} We optimize the asymmetric SupCon loss (eq.~\ref{eqn:supcon_asym}) without an IRD loss. Note that while we do not use buffered samples to learn representations for (a,b), we still need buffered samples to train the downstream linear classifier; for (a,b), we use 200 auxiliary buffered samples to train the classifier (as in (c) and Co$^2$L).

\begin{figure}
\begin{center}
\centerline{\includegraphics[width=0.85\columnwidth]{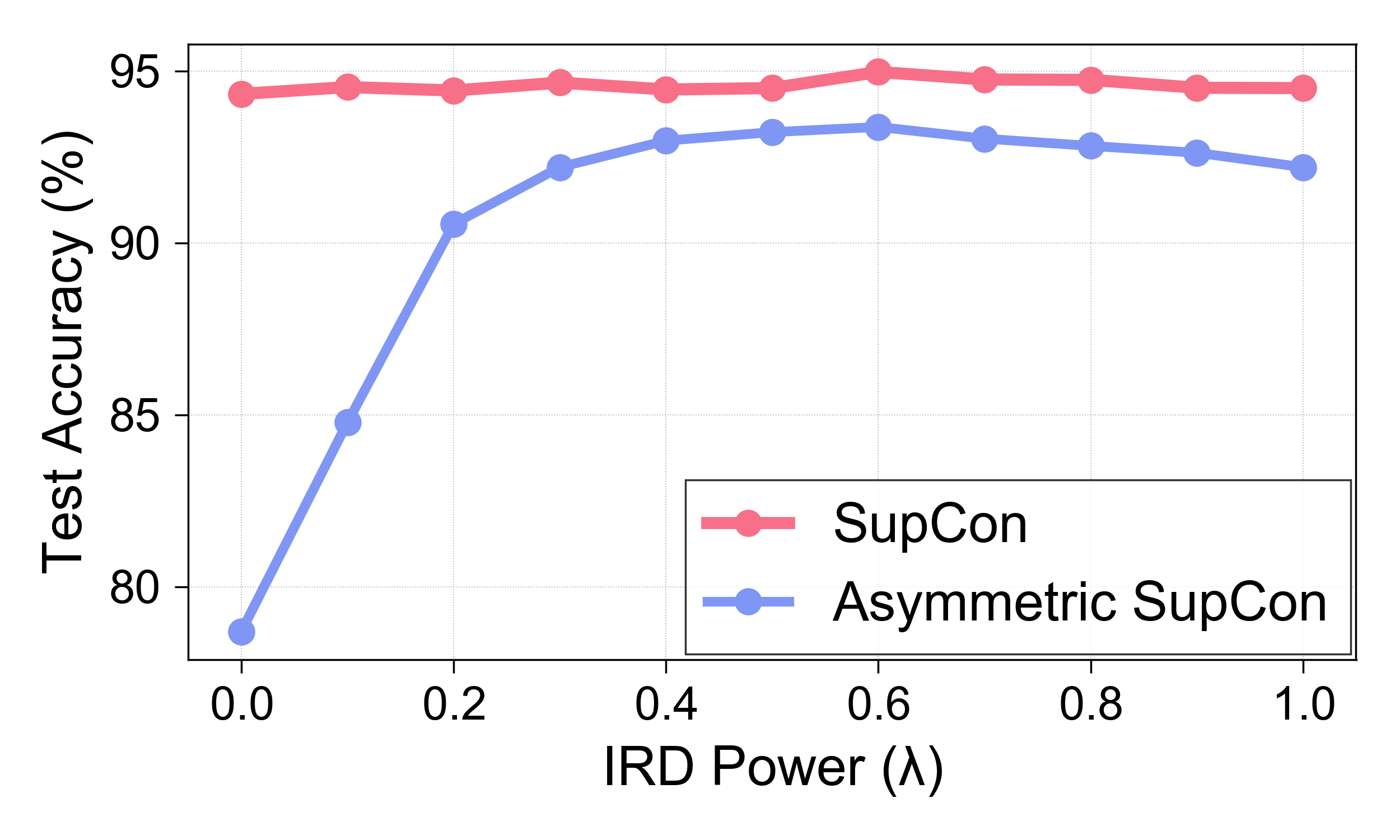}}
\caption{ Performance comparison of original and asymmetric SupCon losses on Seq-CIFAR-10 under the ideal class-IL scenario. Both settings use \text{all} past task samples. Instance-wise Relation Distillation (IRD) effectively closes the performance gap, which indicates IRD successfully retains learned representations without using past samples as positive pairs.}
\label{fig:ird_asym}
\end{center}
\vspace{-2em}
\end{figure}

As shown in Table~\ref{ablation-table}, IRD brings a significant performance gain, with or without the replay buffer. With the replay buffer (rows (c,d)), we observe a 22.40\% relative improvement; without the replay buffer (rows (a,b)), there is a 10.59\% relative improvement. The former is noticeably larger than the latter; we suspect that maintaining the similarity structure of buffered samples (along with current task samples) is essential in preserving learned representations.
We also note that IRD seems to complement the asymmetric SupCon in terms of using buffered samples, leading to a performance boost. To verify this, we consider a synthetic \textit{infinite-buffer} class-IL scenario: all past samples are available throught the training. Under this setup, we train a model with $\mathcal{L}^{\text {sup }}$ and another with $\mathcal{L}^{\text {sup }}_{\text{asym}}$ on Seq-CIFAR-10. As shown in Figure~\ref{fig:ird_asym}, asymmetric SupCon performs relatively poor without using IRD; under this class-balanced setup, not using past task samples as positive pairs only restricts learning. With increasing IRD power, however, the performance gap closes, indicating that IRD complements asymmetric SupCon by helping fully utilize the buffered samples. 
Such trend is also aligned with the results in Table~\ref{ablation-table}; the performance boost from buffered samples--and thus asymmetric SupCon loss--is relatively small without using IRD. This, however, does not necessarily imply that asymmetricity does not bring any benefit, as we will observe in the following ablation study on asymmetric SupCon.



\begin{table}[t]
\setlength{\tabcolsep}{0.2cm}
\centering

\begin{center}
\begin{small}
\begin{tabular}{lccc}
\toprule
& Buffer Size & IRD & Accuracy(\%) \\
\midrule
(a) w/o buffer and IRD & 0 & \color{red}\xmark  & 53.25\stdv{1.70} \\
(b) w/ IRD only & 0 & \color{green}\cmark & 58.89\stdv{2.61} \\
(c) w/ buffer only & 200 & \color{red}\xmark & 53.57\stdv{1.03} \\
(d) Co$^2$L(ours) & 200 & \color{green}\cmark & \textbf{65.57\stdv{1.37}} \\
\bottomrule
\end{tabular}

\end{small}
\end{center}
\caption{Ablation study of Instance-wise Relation Distillation (IRD). We train our model on Seq-CIFAR-10 dataset under class-IL scenario (identical to the setup in Section~\ref{subsection:observation}) with ablated Co$^2$L. IRD brings significant gain with or without replay buffer. All results are averaged over ten independent trials.}
\label{ablation-table}
\end{table}

\vspace{0.2em}
\noindent\textbf{Effectiveness of asymmetric supervised contrastive loss.} 
To verify the effectiveness of asymmetric supervised contrastive loss, we compare two contrastive learning losses, the original SupCon and the asymmetric SupCon, as variants of Co$^2$L  with the identical settings of Section~\ref{subsection:observation}.
As shown in Table~\ref{ablation-table-supcon}, asymmetric SupCon consistently provides gains over all counterparts with the original SupCon.
 
\begin{figure}
\begin{center}
\centerline{\includegraphics[width=0.8\columnwidth]{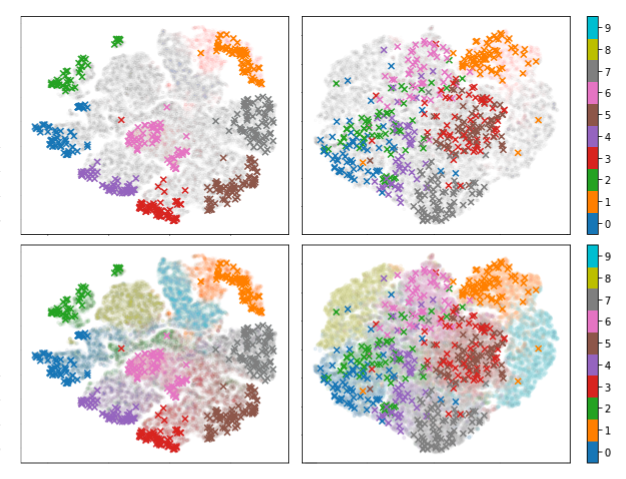}}
\caption{Top: $t$-SNE visualization of features from buffered (colored) and entire (gray) training samples of Seq-CIFAR-10. Bottom: Same as Top, but non-buffered samples are in opaque color instead of gray for a clear illustration of clusters. Left: Buffered samples' features trained with original SupCon are close to the same class samples but distant from different classes. Right: Buffered samples' features trained on asymmetric SupCon are well-spread; buffered samples better represent the entire class sample population.
}
\label{fig:ablation_asym_sup}
\end{center}
\vspace{-3em}
\end{figure}

We also compare the visualizations of encoders' outputs of buffered and entire training samples of the Seq-CIFAR-10 dataset where the encoders are trained in the ablation experiments of Table~\ref{ablation-table-supcon}.
As illustrated in Figure~\ref{fig:ablation_asym_sup}, buffered samples' features trained with original SupCon are close to the same class samples while ones with asymmetric SupCon are well-spread. Since the buffered samples with asymmetric SupCon better represents the entire class sample population, representations trained on asymmetric SupCon show better task performance with linear classifiers. 
Such qualitative results are also well aligned with the motivation of asymmetric SupCon mentioned in Section~\ref{subsection:asymsup} and provide the benefits of asymmetricity.

\begin{table}[t]
\setlength{\tabcolsep}{0.2cm}
\centering
\begin{center}
\begin{small}
\resizebox{\columnwidth}{!}{
\begin{tabular}{ccccccccc}
\toprule
&& \multicolumn{2}{c}{ Seq-CIFAR-10 } & & \multicolumn{2}{c}{ Seq-Tiny-ImageNet } & \\
\cmidrule{3-4}  \cmidrule{6-7}
Buffer  && 200 & 500 & & 200 & 500  &\\
\midrule
$\mathcal{L}^{\text{sup}}$ && 60.49\stdv{0.72} & 68.66\stdv{0.68} & & 13.51\stdv{0.48} & 19.68\stdv{0.62} & \\
$\mathcal{L}^{\text{sup}}_{\text{asym}}$ && \textbf{65.57\stdv{1.37}} & \textbf{74.26\stdv{0.77}} & & \textbf{13.88\stdv{0.40}} & \textbf{20.12\stdv{0.42}} & \\

\bottomrule
\end{tabular}
}
\end{small}
\end{center}

\caption{The effectiveness of asymmetric SupCon loss ($\mathcal{L}^{\text{sup}}_{\text{asym}}$) versus the original SupCon loss ($\mathcal{L}^{\text{sup}}$), combining with the IRD loss. All results are averaged over ten independent trials.}
\label{ablation-table-supcon}

\end{table}


\vspace{-0.3em}
\section{Conclusion}
\label{section:conclusion}
We propose a contrastive continual learning scheme for learning representations under continual learning scenarios. The proposed asymmetric form of contrastive learning loss and the instance-wise relation distillation help model learn and preserve new and past representations and show a better performance over jointly trained baselines on various learning setups. We hope that our work will serve as a good reference to how representation learning for continual learning should be designed.

{\small

\bibliography{references}

\begin{thebibliography}{1}\itemsep=-1pt

\bibitem{SMbuzzega20}
Pietro Buzzega, Matteo Boschini, Angelo Porrello, Davide Abati, and Simone
  Calderara.
\newblock Dark experience for general continual learning: a strong, simple
  baseline.
\newblock In {\em Advances in Neural Information Processing Systems}, 2020.

\bibitem{SMchen20}
Ting Chen, Simon Kornblith, Mohammad Norouzi, and Geoffrey Hinton.
\newblock A simple framework for contrastive learning of visual
  representations.
\newblock In {\em Proceedings of the International Conference on Machine
  Learning}, 2020.

\bibitem{SMfang20}
Zhiyuan Fang, Jianfeng Wang, Lijuan Wang, Lei Zhang, Yezhou Yang, and Zicheng
  Liu.
\newblock {SEED}: Self-supervised distillation for visual representation.
\newblock In {\em International Conference on Learning Representations}, 2021.

\bibitem{SMkhosla20}
Prannay Khosla, Piotr Teterwak, Chen Wang, Aaron Sarna, Yonglong Tian, Phillip
  Isola, Aaron Maschinot, Ce Liu, and Dilip Krishnan.
\newblock Supervised contrastive learning.
\newblock In {\em Advances in Neural Information Processing Systems}, 2020.

\bibitem{SMLoshchilovH17}
Ilya Loshchilov and Frank Hutter.
\newblock {SGDR:} stochastic gradient descent with warm restarts.
\newblock In {\em International Conference on Learning Representations}, 2017.

\bibitem{SMpaszke19}
Adam Paszke, Sam Gross, Francisco Massa, Adam Lerer, James Bradbury, Gregory
  Chanan, Trevor Killeen, Zeming Lin, Natalia Gimelshein, Luca Antiga, Alban
  Desmaison, Andreas Kopf, Edward Yang, Zachary DeVito, Martin Raison, Alykhan
  Tejani, Sasank Chilamkurthy, Benoit Steiner, Lu Fang, Junjie Bai, and Soumith
  Chintala.
\newblock Pytorch: An imperative style, high-performance deep learning library.
\newblock In {\em Advances in Neural Information Processing Systems}.

\end{thebibliography}


\begin{thebibliography}{10}\itemsep=-1pt

\bibitem{Le2015TinyIV}
Stanford 231n.
\newblock Tiny {I}mage{N}et visual recognition challenge.
\newblock \url{https://tiny-imagenet.herokuapp.com}, 2015.

\bibitem{aljundi19}
Rahaf Aljundi, Min Lin, Baptiste Goujaud, and Yoshua Bengio.
\newblock Gradient based sample selection for online continual learning.
\newblock In {\em Advances in Neural Information Processing Systems}, 2019.

\bibitem{arora19}
Sanjeev Arora, Hrishikesh Khandeparkar, Mikhail Khodak, Orestis Plevrakis, and
  Nikunj Saunshi.
\newblock A theoretical analysis of contrastive unsupervised representation
  learning.
\newblock In {\em Proceedings of the International Conference on Machine
  Learning}, 2019.

\bibitem{benjamin19}
Ari~S. Benjamin, David Rolnick, and Konrad~P. K{\"{o}}rding.
\newblock Measuring and regularizing networks in function space.
\newblock In {\em International Conference on Learning Representations}, 2019.

\bibitem{buzzega20}
Pietro Buzzega, Matteo Boschini, Angelo Porrello, Davide Abati, and Simone
  Calderara.
\newblock Dark experience for general continual learning: a strong, simple
  baseline.
\newblock In {\em Advances in Neural Information Processing Systems}, 2020.

\bibitem{chaudhry18}
Arslan Chaudhry, Puneet~K. Dokania, Thalaiyasingam Ajanthan, and Phillip H.~S.
  Torr.
\newblock Riemannian walk for incremental learning: Understanding forgetting
  and intransigence.
\newblock In {\em European Conference on Computer Vision}, 2018.

\bibitem{chaudhry20}
Arslan Chaudhry, Albert Gordo, Puneet~K. Dokania, Philip H.~S. Torr, and David
  Lopez{-}Paz.
\newblock Using hindsight to anchor past knowledge in continual learning.
\newblock In {\em Association for the Advancement of Artificial Intelligence},
  2020.

\bibitem{chaudhry19}
Arslan Chaudhry, Marc'Aurelio Ranzato, Marcus Rohrbach, and Mohamed Elhoseiny.
\newblock Efficient lifelong learning with {A-GEM}.
\newblock In {\em International Conference on Learning Representations}, 2019.

\bibitem{Chawla02}
Nitesh~V. Chawla, Kevin~W. Bowyer, Lawrence~O. Hall, and W.~Philip Kegelmeyer.
\newblock {SMOTE}: Synthetic minority over-sampling technique.
\newblock {\em Journal of artificial intelligence research}, 2002.

\bibitem{chen20}
Ting Chen, Simon Kornblith, Mohammad Norouzi, and Geoffrey Hinton.
\newblock A simple framework for contrastive learning of visual
  representations.
\newblock In {\em Proceedings of the International Conference on Machine
  Learning}, 2020.

\bibitem{Chen19}
Ting Chen, Xiaohua Zhai, Marvin Ritter, Mario Lucic, and Neil Houlsby.
\newblock Self-supervised gans via auxiliary rotation loss.
\newblock In {\em Proceedings of the IEEE/CVF Conference on Computer Vision and
  Pattern Recognition}, 2019.

\bibitem{chen2020exploring}
Xinlei Chen and Kaiming He.
\newblock Exploring simple siamese representation learning, 2020.

\bibitem{fang20}
Zhiyuan Fang, Jianfeng Wang, Lijuan Wang, Lei Zhang, Yezhou Yang, and Zicheng
  Liu.
\newblock {SEED}: Self-supervised distillation for visual representation.
\newblock In {\em International Conference on Learning Representations}, 2021.

\bibitem{finn17}
Chelsea Finn, Pieter Abbeel, and Sergey Levine.
\newblock Model-agnostic meta-learning for fast adaptation of deep networks.
\newblock In {\em Proceedings of the International Conference on Machine
  Learning}, 2017.

\bibitem{grill2020bootstrap}
Jean-Bastien Grill, Florian Strub, Florent Altché, Corentin Tallec, Pierre~H.
  Richemond, Elena Buchatskaya, Carl Doersch, Bernardo~Avila Pires,
  Zhaohan~Daniel Guo, Mohammad~Gheshlaghi Azar, Bilal Piot, Koray Kavukcuoglu,
  Rémi Munos, and Michal Valko.
\newblock Bootstrap your own latent: A new approach to self-supervised
  learning, 2020.

\bibitem{gupta20}
Gunshi Gupta, Karmesh Yadav, and Liam Paull.
\newblock Look-ahead meta learning for continual learning.
\newblock In {\em Advances in Neural Information Processing Systems}, 2020.

\bibitem{gutmann10}
Michael Gutmann and Aapo Hyvärinen.
\newblock Noise-contrastive estimation: A new estimation principle for
  unnormalized statistical models.
\newblock In {\em Proceedings of the International Conference on Machine
  Learning}, 2010.

\bibitem{hadsell06}
Raia Hadsell, Sumit Chopra, and Yann LeCun.
\newblock Dimensionality reduction by learning an invariant mapping.
\newblock In {\em Proceedings of the IEEE/CVF Conference on Computer Vision and
  Pattern Recognition}, 2006.

\bibitem{han05}
Hui Han, Wen-Yuan Wang, and Bing-Huan Mao.
\newblock Borderline-{SMOTE}: A new over-sampling method in imbalanced data
  sets learning.
\newblock In {\em International Conference on Intelligent Computing}, 2005.

\bibitem{he20}
Kaiming He, Haoqi fan, Yuxin Wu, Saining Xie, and Ross Girshick.
\newblock Momentum contrast for unsupervised visual representation learning.
\newblock In {\em Proceedings of the IEEE/CVF Conference on Computer Vision and
  Pattern Recognition}, 2020.

\bibitem{He2015}
Kaiming He, Xiangyu Zhang, Shaoqing Ren, and Jian Sun.
\newblock Deep residual learning for image recognition.
\newblock {\em arXiv preprint arXiv:1512.03385}, 2015.

\bibitem{javed19}
Khurram Javed and Martha White.
\newblock Meta-learning representations for continual learning.
\newblock In {\em Advances in Neural Information Processing Systems}, 2019.

\bibitem{khosla20}
Prannay Khosla, Piotr Teterwak, Chen Wang, Aaron Sarna, Yonglong Tian, Phillip
  Isola, Aaron Maschinot, Ce Liu, and Dilip Krishnan.
\newblock Supervised contrastive learning.
\newblock In {\em Advances in Neural Information Processing Systems}, 2020.

\bibitem{kirkpatrick17}
James Kirkpatrick, Razvan Pascanu, Neil~C. Rabinowitz, Joel Veness, Guillaume
  Desjardins, Andrei~A. Rusu, Kieran Milan, John Quan, Tiago Ramalho, Agnieszka
  Grabska-Barwinska, Demis Hassabis, Claudia Clopath, Dharshan Kumaran, and
  Raia Hadsell.
\newblock Overcoming catastrophic forgetting in neural networks.
\newblock {\em Proceedings of the National Academy of Sciences of the United
  States of America}, 2017.

\bibitem{krizhevsky2009learning}
Alex Krizhevsky, Geoffrey Hinton, et~al.
\newblock Learning multiple layers of features from tiny images.
\newblock Technical report, University of Toronto, 2009.

\bibitem{lecun98}
Yann LeCun, Léon Bottou, Yoshua Bengio, and Patrick Haffner.
\newblock Gradient-based learning applied to document recognition.
\newblock In {\em Proceedings of the IEEE}, 1998.

\bibitem{Li16}
Zhizhong Li and Derek Hoiem.
\newblock Learning without forgetting.
\newblock In {\em European Conference on Computer Vision}, 2016.

\bibitem{lopezpaz17}
David Lopez-Paz and Marc'Aurelio Ranzato.
\newblock Gradient episodic memory for continual learning.
\newblock In {\em International Conference on Learning Representations}, 2017.

\bibitem{mallya18}
Arun Mallya and Svetlana Lazebnik.
\newblock Pack{N}et: Adding multiple tasks to a single network by iterative
  pruning.
\newblock In {\em Proceedings of the IEEE/CVF Conference on Computer Vision and
  Pattern Recognition}, 2018.

\bibitem{mccloskey89}
Michael McCloskey and Neal~J. Cohen.
\newblock Catastrophic interference in connectionist networks: The sequential
  learning problem.
\newblock {\em Psychology of Learning and Motivation}, 1989.

\bibitem{Rebuffi17}
Sylvestre-Alvise Rebuffi, Alexander Kolesnikov, Georg Sperl, and Christoph~H.
  Lampert.
\newblock icarl: Incremental classifier and representation learning.
\newblock In {\em Proceedings of the IEEE/CVF Conference on Computer Vision and
  Pattern Recognition}, 2017.

\bibitem{riemer19}
Matthew Riemer, Ignacio Cases, Robert Ajemian, Miao Liu, Irina Rish, Yuhai Tu,
  and Gerald Tesauro.
\newblock Learning to learn without forgetting by maximizing transfer and
  minimizing interference.
\newblock In {\em International Conference on Learning Representations}, 2019.

\bibitem{robins95}
Anthony Robins.
\newblock Catastrophic forgetting, rehearsal, and pseudorehearsal.
\newblock {\em Connection Science}, 1995.

\bibitem{robinson21}
Joshua~David Robinson, Ching-Yao Chuang, Suvrit Sra, and Stefanie Jegelka.
\newblock Contrastive learning with hard negative samples.
\newblock In {\em International Conference on Learning Representations}, 2021.

\bibitem{ILSVRC15}
Olga Russakovsky, Jia Deng, Hao Su, Jonathan Krause, Sanjeev Satheesh, Sean Ma,
  Zhiheng Huang, Andrej Karpathy, Aditya Khosla, Michael Bernstein,
  Alexander~C. Berg, and Li Fei-Fei.
\newblock Image{N}et large scale visual recognition challenge.
\newblock {\em International Journal of Computer Vision}, 2015.

\bibitem{rusu16}
Andrei~A. Rusu, Neil~C. Rabinowitz, Guillaume Desjardins, Hubert Soyer, James
  Kirkpatrick, Koray Kavukcuoglu, Razvan Pascanu, and Raia Hadsell.
\newblock Progressive neural networks.
\newblock {\em arXiv preprint 1606.04671}, 2016.

\bibitem{serra18}
Joan Serra, Didac Suris, Marius Miron, and Alexandros Karatzoglou.
\newblock Overcoming catastrophic forgetting with hard attention to the task.
\newblock In {\em Proceedings of the International Conference on Machine
  Learning}, 2018.

\bibitem{tian19}
Yonglong Tian, Dilip Krishnan, and Phillip Isola.
\newblock Contrastive multiview coding.
\newblock {\em arXiv preprint arXiv:1906.05849}, 2019.

\bibitem{vandeven2019three}
Gido~M. van~de Ven and Andreas~S Tolias.
\newblock Three scenarios for continual learning.
\newblock {\em arXiv preprint arXiv:1904.07734}, 2019.

\bibitem{oord18}
A{\"{a}}ron van~den Oord, Yazhe Li, and Oriol Vinyals.
\newblock Representation learning with contrastive predictive coding.
\newblock {\em arXiv preprint arXiv:1807.03748}, 2018.

\end{thebibliography}
}

\clearpage
\appendix
\label{appendix}

\section{Training Details}
\label{appendix:training_details}
\subsection{Augmentation}
We follow the data augmentation scheme introduced in \citeSM{SMchen20} for representation learning and linear evaluation. 
We describe the default set of augmentations following the PyTorch \citeSM{SMpaszke19} notations in what follows. 

\begin{itemize}[leftmargin=*,topsep=0pt,itemsep=-0.5ex,partopsep=1ex,parsep=1ex]
\item \texttt{RandomResizedCrop.} We crop Seq-CIFAR-10, Tiny-ImageNet, and R-MNIST datasets with the scale in $[0.2, 1.0]$, $[0.1, 1.0]$, and $[0.7, 1.0]$, respectively. The cropped images are resized to $32 \times 32$ for Seq-CIFAR-10, $64 \times 64$ for Tiny-ImageNet, and $28 \times 28$ for R-MNIST.
\item \texttt{RandomHorizontalFlip.} Images are flipped horizontally with probability $0.5$.
\item \texttt{ColorJitter.} The maximum strengths of \{brightness, contrast, saturation, hue\} are \{0.4, 0.4, 0.4, 0.1\} with probability $0.8$.
\item \texttt{RandomGrayScale.} Images are grayscaled with probability $0.2$.
\item \texttt{GaussianBlur.} For the Tiny-ImageNet dataset, blur augmentation is applied with Gaussian kernel. Kernel size is $7 \times 7$ and the standard deviation is randomly drawn from $[0.1, 2.0]$. This operation is randomly applied with probability $0.5$.
\end{itemize}

\subsection{Architecture}
For Seq-CIFAR-10 and Tiny-ImageNet datasets, we use ResNet-18 (not pretrained) as a base encoder for representation learning followed by a 2-layer projection MLP which maps representations to a 128-dimensional latent space \citeSM{SMkhosla20}. The hidden layer of projection MLP consists of 512 hidden units.

For R-MNIST, we use two convolutional layers and one fully connected layer for the base encoder. We use 20 and 50 filters with $5 \times 5$ kernel and stride $1$ for two convolutional layers, respectively. Each feature map is followed by a max pooling operation with stride $2$. 
The base encoder for R-MNIST is also followed by a 2-layer projection MLP for representation learning.
The output dimensions of the last fully connected layer of encoder and the following 2-layer MLP's all hidden/output neuron sizes are equally 500.

\subsection{Hyperparameter}
\begin{table}[t]
\centering
\begin{small}{
\resizebox{0.9\columnwidth}{!}{
\begin{tabular}{lcl}
\toprule
\textit{Method} & \textit{Buffer} & \textit{Parameters} \\
\midrule
\multicolumn{3}{c}{ \textbf{R-MNIST} } \\
\midrule
ER & 200 & $\eta$: 0.1 \\
& 500 & $\eta$: 0.1 \\
GEM & 200 & $\eta$: 0.1, $\gamma$: 0.5 \\
& 500 & $\eta$: 0.3, $\gamma$: 0.5 \\
A-GEM & 200 & $\eta$: 0.1 \\
& 500 & $\eta$: 0.1 \\
FDR & 200 & $\eta$: 0.1, $\alpha$: 1.0 \\
& 500 & $\eta$: 0.2, $\alpha$: 0.3 \\
GSS & 200 & $\eta$: 0.2, \textit{gmbs}: 128, \textit{nb}: 1 \\
& 500 & $\eta$: 0.2, \textit{gmbs}: 128, \textit{nb}: 1 \\
HAL & 200 & $\eta$: 0.03, $\lambda$:0.1, $\beta$: 0.3, $\alpha$:0.1 \\
& 500 & $\eta$: 0.03, $\lambda$:0.1, $\beta$: 0.5, $\alpha$:0.1 \\
DER & 200 & $\eta$: 0.1, $\alpha$: 0.5 \\
& 500 & $\eta$: 0.1, $\alpha$: 0.5 \\
DER++ & 200 & $\eta$: 0.1, $\alpha$: 1.0, $\beta$: 0.5 \\
& 500 & $\eta$: 0.2, $\alpha$: 1.0, $\beta$: 1.0 \\
Co$^2$L & 200 & \multirow{2}{*}{$\eta$: 0.01, $\tau$:0.1, $\kappa$: 0.2, $\kappa^*$: 0.01, \textit{epoch}: 20 } \\
& 500 & \\
\midrule
\multicolumn{3}{c}{ \textbf{Seq-CIFAR-10} } \\
\midrule
Co$^2$L & 200 & \multirow{2}{*}{$\eta$: 0.5, $\tau$:0.5, $\kappa$: 0.2, $\kappa^*$: 0.01, \textit{epoch}: 100 } \\
& 500 & \\
\midrule
\multicolumn{3}{c}{ \textbf{Seq-Tiny-ImageNet} } \\
\midrule
Co$^2$L & 200 & \multirow{2}{*}{$\eta$: 0.1, $\tau$:0.5, $\kappa$: 0.1, $\kappa^*$: 0.1, \textit{epoch}: 50 } \\
& 500 & \\
\bottomrule
\end{tabular}
}
}
\end{small}
\caption{Hyperparameters chosen in our experiments}
\label{chosen-param-table}
\end{table} 
\begin{table}[t]
\centering
\begin{small}{
\resizebox{0.9\columnwidth}{!}{
\begin{tabular}{ccc}
\toprule
\textit{Dataset} & \textit{Parameter} & \textit{Values} \\
\midrule
\multirow{7}{*}{\textbf{Seq-CIFAR-10}} & $\eta$ & \{0.1, 0.5, 1.0\} \\
& $\tau$ & \{0.1, 0.5, 1.0\} \\
& $\kappa$ & \{0.1, 0.2\} \\
& $\kappa^*$ & \{0.01, 0.05, 0.1\} \\
& $E_{0}$ & \{500\} \\
& $E_{t>0}$ & \{50, 100\} \\
& \textit{bsz} & \{256, 512, 1024\} \\
\midrule
\multirow{7}{*}{\textbf{Seq-Tiny-ImageNet}} & $\eta$ & \{0.1, 0.5, 1.0\} \\
& $\tau$ & \{0.1, 0.5, 1.0\} \\
& $\kappa$ & \{0.1, 0.2\} \\
& $\kappa^*$ & \{0.01, 0.05, 0.1\} \\
& $E_{0}$ & \{500\} \\
& $E_{t>0}$ & \{50, 100\} \\
& \textit{bsz} & \{256, 512, 1024\} \\
\midrule
\multirow{7}{*}{\textbf{R-MNIST}} & $\eta$ & \{0.01, 0.05, 0.1\} \\
& $\tau$ & \{0.1, 0.5, 1.0\} \\
& $\kappa$ & \{0.1, 0.2\} \\
& $\kappa^*$ & \{0.01, 0.05, 0.1\} \\
& $E_{0}$ & \{100\} \\
& $E_{t>0}$ & \{10, 20\} \\
& \textit{bsz} & \{256, 512, 1024\} \\
\bottomrule
\end{tabular}
}
}
\end{small}
\caption{Hyperparameter space for Co$^2$L}
\label{param-search-table}
\end{table}

The hyperparameters for Section~\ref{section:experiment} are selected by performing a grid search using the validation set consisting of randomly drawn 10\% of the training samples, and chosen hyperparameters are given in Table~\ref{chosen-param-table}. We consider following hyperparameters for Co$^2$L: learning rate ($\eta$), batch size (\textit{bsz}), temperature for asymmetric supervised contrastive learning loss ($\tau$), temperatures for instance-wise relation distillation loss ($\kappa, \kappa^*$), and the number of epochs of $t$-th task ($E_t$). The hyperparameter search space for Co$^2$L on benchmark datasets are provided in Table~\ref{param-search-table}. 
In a combined grid search for Class-IL and Task-IL, we select the best hyperparameters that achieve the highest final accuracy averaged over both settings.
For R-MNIST, we conduct a grid search for all baselines since the architecture for R-MNIST changes. We follow the hyperparameter search space (and its notations) for R-MNIST given in \citeSM{SMbuzzega20}. For all experiments of Co$^2$L, we use distillation power ($\lambda$ in eq.~\ref{eqn:total-loss}) as $1.0$.

\subsection{Training Details for Co$^2$L}
For representation learning, we use a linear warmup for the first 10 epochs and decay the learning rate with the cosine decay schedule \citeSM{SMLoshchilovH17}. The learning rate scheduling is restarted at every task is introduced. We use SGD with momentum $0.9$ and weight decay $0.0001$ for all experiments.

For linear evaluation, we train a linear classifier for 100 epochs using SGD with momentum $0.9$ and no weight decay. We decay the learning rate exponentially at $60$, $75$, and $90$ epoch with decay rate 0.2. We use \{1.0, 0.1, 1.0\} learning rate for \{Seq-CIFAR-10, Seq-Tiny-ImageNet, R-MNIST\}.

\clearpage

\begin{table*}[t]
\setlength{\tabcolsep}{0.2cm}
\centering
\begin{center}
\begin{small}
\resizebox{0.75\textwidth}{!}{
\begin{tabular}{clccccccccc}
\toprule
\multirow{2}{*}{ Buffer } & \multirow{2}{*}{ Objective } & \multirow{2}{*}{ Space } & & \multicolumn{2}{c}{ \textbf{Seq-CIFAR-10} } & & \multicolumn{2}{c}{ \textbf{Seq-Tiny-ImageNet} } &  \\
\cmidrule{5-6} \cmidrule{8-9}
&  & & & Class-IL & Task-IL  & & Class-IL & Task-IL & \\
\midrule
\multirow{4}{*}{ 200 } 
& $\mathcal{L}^{\text{asym}}_{\text{sup}} + \mathcal{L}^{\text{SEED}}$ \citeSM{SMfang20}
& \multirow{2}{*}{ Embedding } 
& & 53.42\stdv{1.07} & 85.79\stdv{0.91} 
& & 9.23\stdv{0.65} & 27.02\stdv{1.70}
&  \\
& $\mathcal{L}^{\text{asym}}_{\text{sup}} + \mathcal{L}^{\text{MSE}}_{\text{embedding}}$
& 
& & 56.31\stdv{2.30} & 86.12\stdv{0.94} 
& & 11.03\stdv{0.26} & 33.15\stdv{0.55}
&  \\
& $\mathcal{L}^{\text{asym}}_{\text{sup}} + \mathcal{L}^{\text{MSE}}_{\text{projection}}$ 
& \multirow{2}{*}{ Projection } 
& & 53.10\stdv{1.52} & 85.05\stdv{0.95} 
& & 11.45\stdv{0.33} & 34.38\stdv{0.66}
&  \\
& $\mathcal{L}^{\text{asym}}_{\text{sup}} + \mathcal{L}^{\text{IRD}}$ (ours)
& 
& & \textbf{65.57\stdv{1.37}}& \textbf{93.43\stdv{0.78}} 
& & \textbf{13.88\stdv{0.40}} & \textbf{42.37\stdv{0.74}}
& \\

\midrule
\multirow{4}{*}{ 500 } 
& $\mathcal{L}^{\text{asym}}_{\text{sup}} + \mathcal{L}^{\text{SEED}}$ \citeSM{SMfang20}
& \multirow{2}{*}{ Embedding } 
& & 61.65\stdv{3.24} & 88.40\stdv{2.44} 
& & 12.04\stdv{0.40} & 34.91\stdv{0.57} 
&\\
& $\mathcal{L}^{\text{asym}}_{\text{sup}} + \mathcal{L}^{\text{MSE}}_{\text{embedding}}$
& 
& & 62.83\stdv{2.92} & 88.63\stdv{2.05} 
& & 14.89\stdv{0.40} & 42.25\stdv{0.51}
&  \\
& $\mathcal{L}^{\text{asym}}_{\text{sup}} + \mathcal{L}^{\text{MSE}}_{\text{projection}}$ 
& \multirow{2}{*}{ Projection } 
& & 57.47\stdv{1.07} & 86.29\stdv{0.31} 
& & 14.73\stdv{0.39} & 41.85\stdv{1.22}
&  \\
& $\mathcal{L}^{\text{asym}}_{\text{sup}} + \mathcal{L}^{\text{IRD}}$ (ours)
& 
& & \textbf{74.26\stdv{0.77}}& \textbf{95.90\stdv{0.26}} 
& & \textbf{20.12\stdv{0.42}} & \textbf{53.04\stdv{0.69}}
&\\
\bottomrule
\end{tabular}
}
\end{small}
\end{center}
\caption{Classification accuracies for Seq-CIFAR-10 and Seq-Tiny-ImageNet on our algorithm and three alternatives. All results are averaged over ten independent trials.
}
\label{seed-mse-table}
\end{table*}

\section{Experiments on IRD Alternatives}
We propose Co$^2$L that learns representations and preserves learned representations using $\mathcal{L}^\text{asym}_\text{sup}$ and $\mathcal{L}^\text{IRD}$, respectively. In this section, we explore alternatives for IRD loss to preserve learned representations, and verify its effectiveness. More specifically, we consider following baselines.

\noindent\textbf{Embedding distillation}. 
IRD can be viewed as distilling the representations from the past self, similar to how SEED \citeSM{SMfang20} distills the representation from the teacher model to the student model.
However, there is a slight difference: IRD distills the instance-wise similarity of the outputs from the joint encoder-projector, where the projector is introduced for contrastive learning. SEED directly distills the output of the encoder.
Specifically, for each sample $\tilde{x}_i$ in a batch $\mathcal{B}$, the similarity score with respect to an encoder $f_\vartheta$ is defined as:
\begin{align}
\label{eqn:simvec_seed}
    \mathbf{p}\left(\tilde{\mathbf{x}}_{i} ; \vartheta, \gamma\right)=\left[p_{i,1}, \ldots, p_{i,2N}\right],
\end{align}
where $p_{i,j}$ denotes the normalized similarity
\begin{align}
p_{i,j}&=\frac{\exp \left(\mathbf{z}_{i} \cdot \mathbf{z}_{j} / \gamma\right)}{\sum_{k \neq i}^{2N} \exp \left(\mathbf{z}_{i} \cdot \mathbf{z}_{k} / \gamma\right)},
\end{align}
and $\mathbf{z}_{i}$ denotes the normalized feature vector representations of $\tilde{x}_i$ from the encoder $f_\vartheta$, \ie, $\mathbf{z}_{i} = f_{\vartheta}(\tilde{x}_i) / \|f_{\vartheta}(\tilde{x}_i)\|_{2}$.
Denoting the parameters of the teacher/student encoder and temperature as $\vartheta^\text{T}, \vartheta^\text{S}$ and $\gamma^\text{T}, \gamma^\text{S}$, the SEED loss is defined as 
\begin{align}
\label{eqn:SEED}
    \mathcal{L}^{\text {SEED }}  &=\sum_{i=1}^{2N}-\mathbf{p}\left(\tilde{\mathbf{x}}_{i} ; \vartheta^{\text{T}}, \gamma^\text{T}\right) \cdot \log \mathbf{p}\left(\tilde{\mathbf{x}}_{i} ; \vartheta^\text{S}, \gamma^\text{S}\right)
\end{align}

\noindent\textbf{Logit matching}. Buzzega \etal \citeSM{SMbuzzega20} shows matching the logit, \ie, pre-softmax outputs, of the past and current model is effective for mitigating forgetting. Similarly, we replace IRD loss with the one that directly matches representation maps. Specifically, for each sample $\tilde{x}_i$ in batch $\mathcal{B}$, two types of matching loss are defined as
\begin{gather}
\mathcal{L}^{\text{MSE}}_{\text{embedding}} = \frac{1}{2N} \sum_{i=1}^{2N} \big(f_\vartheta(\tilde{x}_i) - f_{\vartheta^*}(\tilde{x}_i)\big)^2 \\
\mathcal{L}^{\text{MSE}}_{\text{projection}} = \frac{1}{2N} \sum_{i=1}^{2N} \big((g \circ f)_{\psi}(\tilde{x}_i) - (g \circ f)_{\psi^*}(\tilde{x}_i)\big)^2
\end{gather}
where $f_\vartheta$ is encoder and $(g \circ f)_{\psi}$ is the feature map which maps augmented batch to an unnormalized $d$-dimensional Euclidean sphere. The difference between $\mathcal{L}^{\text{MSE}}_{\text{embedding}}$ and $\mathcal{L}^{\text{MSE}}_{\text{projection}}$ is the choice of representation maps to be matched; one defined on embedding space and the other defined on the projection space.

As shown in Table~\ref{seed-mse-table}, we find that distilling the projector output (and thereby applying both $\mathcal{L}^\text{IRD}$ and $\mathcal{L}^\text{sup}_\text{asym}$ at the same layer) significantly outperforms distilling at the encoder output ($\mathcal{L}^\text{SEED}$, $\mathcal{L}^{\text{MSE}}_{\text{embedding}}$) and the projection output ($\mathcal{L}^{\text{MSE}}_{\text{projection}}$). 
Since we learn representations continually in constrastive learning schemes, where similarity is defined on a unit $d$-dimensional Euclidean sphere, regulating the relation drifts in the projection space can be more effective to preserve learned representations than other alternatives.

\newpage
{\small
\bibliographystyleSM{ieee_fullname}
\bibliographySM{references_appendix}
}

\end{document}